\documentclass[]{IEEEtran}
\usepackage{cite}
\usepackage[pdftex]{graphicx}
\usepackage{amsmath}
\usepackage{algorithm}
\usepackage{algorithmic}

\usepackage{multirow}
\usepackage{booktabs}
\usepackage{amssymb}
\usepackage{float}
\usepackage{stfloats}
\usepackage{booktabs}
\usepackage{graphicx}
\usepackage{multirow}
\usepackage{amsmath}
\usepackage{algorithm}
\usepackage{algorithmic}
\usepackage{lineno}
\usepackage{amsfonts,amssymb}
\usepackage{bbding}
\usepackage{color}
\usepackage{hyperref}
\usepackage{makecell}
\usepackage{orcidlink} 
\usepackage{colortbl}
\usepackage{graphicx} 	
\usepackage{subfigure}	
\usepackage{algorithm}  
\usepackage{hyperref}
\hypersetup{hidelinks,
	colorlinks=true,
	allcolors=blue,
	pdfstartview=Fit,
	breaklinks=true}

\usepackage[numbers,sort&compress]{natbib}

\usepackage{float}
\usepackage{stfloats}

\usepackage{hyperref}

\begin{document}
\title{ Physical Depth-aware Early Accident Anticipation: A Multi-dimensional Visual Feature Fusion Framework}

\author{Hongpu~Huang\thanks{Hongpu~Huang\textsuperscript{†} and Wei~Zhou\textsuperscript{†} contributed equally to this work as co-first authors.}\textsuperscript{†},\orcidlink{0009-0003-0210-6589},  
    Wei~Zhou\textsuperscript{†},\orcidlink{0000-0003-3225-0576}\IEEEmembership{Graduate Student Member,~IEEE},  
    and Chen~Wang,\orcidlink{0000-0003-4573-9047}\IEEEmembership{Member,~IEEE}

  \thanks{This research is supported by the National Key Research and Development Program of China (2023YFE0106800) and the Science Fund for Distinguished Young Scholars of Jiangsu Province (BK20231531).

Hongpu Huang\textsuperscript{†}  and Chen Wang (\textit{Corresponding Author}) are with the School of Transportation, Southeast University, Nanjing 211189, China. Wei Zhou\textsuperscript{†} is with the School of Transportation, Southeast University, Nanjing 211189, China,  and also with the Department of Industrial and Systems Engineering, The Hong Kong Polytechnic University, Hong Kong, SAR, China. (e-mail: 220223065@seu.edu.cn; vvgod@seu.edu.cn; chen\_david\_wang@seu.edu.cn).}
  }

\markboth{IEEE TRANSACTIONS ON INTELLIGENT VEHICLES}
{Shell \MakeLowercase{\textit{et al.}}: Bare Demo of IEEEtran.cls for IEEE Journals}
%

\maketitle

\begin{abstract}
Early accident anticipation from dashcam videos is a highly desirable yet challenging task for improving the safety of intelligent vehicles. Existing advanced accident anticipation approaches commonly model the interaction among traffic agents (e.g., vehicles, pedestrians, etc.) in the coarse 2D image space, which may not adequately capture their true positions and interactions. To address this limitation, we propose a physical depth-aware learning framework that incorporates the monocular depth features generated by a large model named Depth-Anything to introduce more fine-grained spatial 3D information. Furthermore, the proposed framework also integrates visual interaction features and visual dynamic features from traffic scenes to provide a more comprehensive perception towards the scenes. Based on these multi-dimensional visual features, the framework captures early indicators of accidents through the analysis of interaction relationships between objects in sequential frames. Additionally, the proposed framework introduces a reconstruction adjacency matrix for key traffic participants that are occluded, mitigating the impact of occluded objects on graph learning and maintaining the spatio-temporal continuity. Experimental results on public datasets show that the proposed framework attains state-of-the-art performance, highlighting the effectiveness of incorporating visual depth features and the superiority of the proposed framework.

\end{abstract}


%

\IEEEpeerreviewmaketitle

\section{INTRODUCTION}
\label{ch1}

%
%
%
%

\begin{figure}[htbp]  
\centering  
\includegraphics[width=0.5\textwidth]{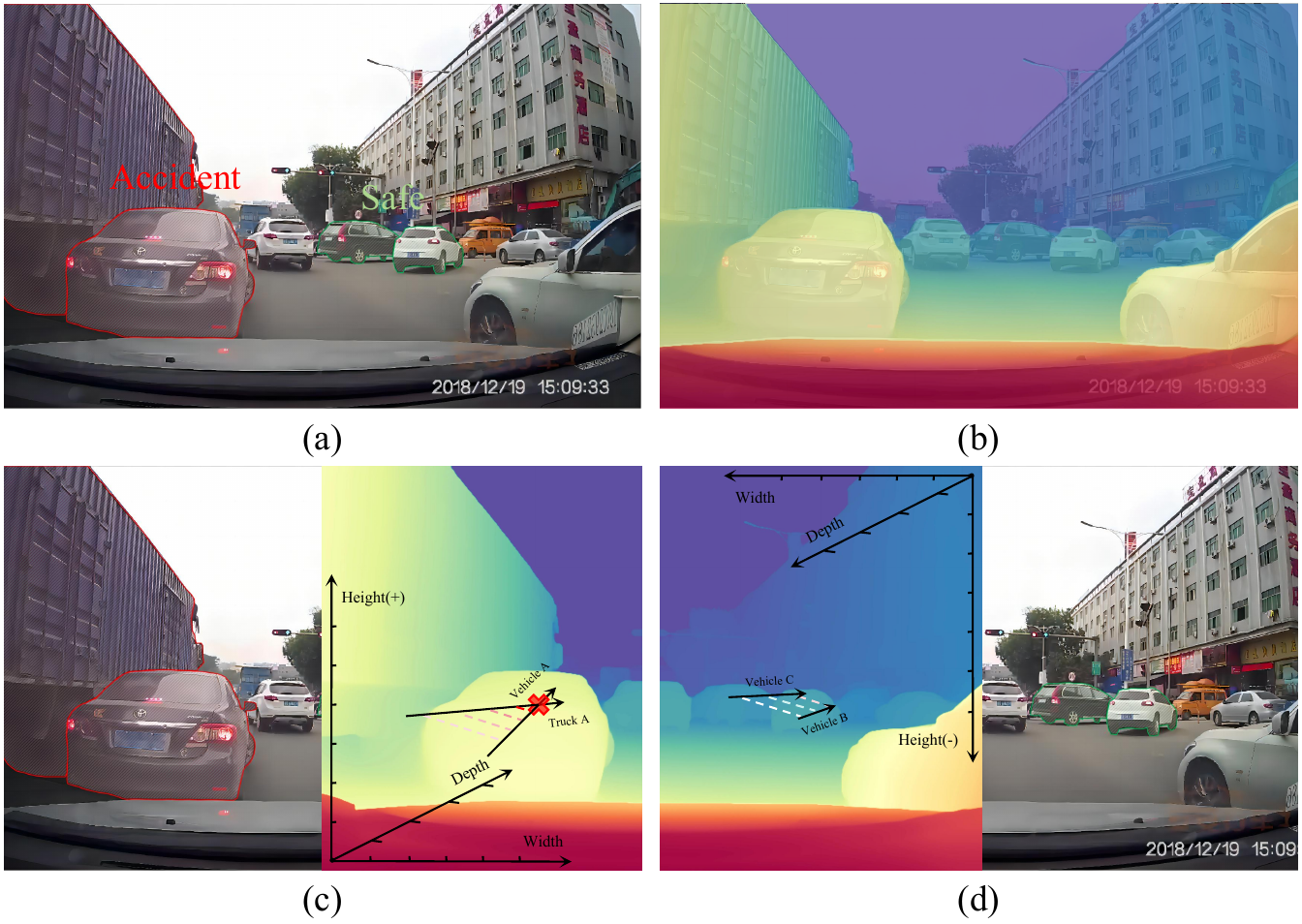}  
\caption{(a) A key frame depicting a traffic accident, where the red vehicles are involved in the accident, while the green vehicles are overlapping in position but not involved in the accident; (b) The visualized depth map of the key frame; (c) The vehicles involved in the accident intersect in the depth dimension. (d) The vehicles not involved in the accident overlap in position but do not intersect in the depth dimension.}  
\label{fig1}
\end{figure}

\IEEEPARstart{A}{s}SISTIVE driving technologies have received considerable attention to enhance road traffic safety and reduce the number of accidents. Despite the continuous emergence of new vehicles equipped with advanced driver assistance systems (ADAS) such as Adaptive Cruise Control (ACC) and Automatic Emergency Braking (AEB) in recent years \cite{xiao2010comprehensive,tan2020automatic}, the number of motor vehicle accidents remains increasingly high. According to the China Statistical Yearbook (NBSC) \cite{NBSC2022}, China alone witnessed 233,729 motor vehicle accidents in 2021, which results in billions of economic losses. If the accidents can be predicted in advance through the widespread deployment of dashcams, even just a few seconds earlier, it can provide human drivers and ADAS with ample response time. This would significantly improve the drivers’ safety in the complex vehicle driving environment.

Road regulations and safety features are often based on human visual requirements \cite{sainju2020mapping,zhou2024vision,zhou2024teaching}. Visual-based solutions to perceive the road environment closely align with the experiences of human drivers \cite{fang2023vision}. Due to noise interference in radar data, vision-based solutions sometimes outperform radar-vision fusion approaches in certain scenarios \cite{lin2020depth}. Visual solutions for early accident warning rely on image data, while dashcams provide valuable video recordings that capture key information before and after accidents. By effectively identifying and extracting key factors that contribute to accidents, researchers can enhance the early accident prediction accuracy and better prepare for potential subsequent accidents \cite{karim2022dynamic}. Consequently, the development of methods for extracting and capturing early warning signs from dashcam video data has become a pressing research priority.

Existing advanced research \cite{suzuki2018anticipating,fatima2021global,malawade2022spatiotemporal,wang2023gsc,yao2015describing,chan2017anticipating,bao2020uncertainty,zhou2023appearance,zhou2023monitoring} oftentimes developed deep learning networks that learn spatiotemporal features in traffic scenarios for early accident anticipation. In general, these methods first construct agent features such as appearance features, motion trajectories, etc., and utilize attention allocation \cite{suzuki2018anticipating,fatima2021global} or graph computation \cite{malawade2022spatiotemporal,wang2023gsc,zhou2023pedestrian} to achieve spatial interaction between agents. Subsequently, they typically utilize dynamic temporal attention modules \cite{yao2015describing,chan2017anticipating,bao2020uncertainty} or recurrent neural networks \cite{godard2017unsupervised} to further capture temporal features in dynamic traffic scenarios. Although remarkable results have been achieved, their modeling of interactions between agents is based on the 2D image space that incorporate implicit depth information. Due to the existence of perspective distortion, the 2D image space lacks depth information, which may make it difficult to accurately measure the distance and position between agents \cite{cho2014learning,fu2018deep,godard2019digging}. As shown in Fig.\ref{fig1}, the inclusion of depth information has the potential to enhance the model's ability to accurately perceive the distance of traffic scenes, as well as bypass false alarms where traffic agents appear to overlap in the image but do not overlap in depth.

Recently, a large model targeting the task of depth estimation, named Depth-anything \cite{yang2024depth}, is proposed to produce explicit depth information for various images, through learning on a massive amount of data. Inspired by this, this paper combines the powerful monocular depth estimation capability of Depth-Anything to construct a physical depth-aware learning framework. Specifically, the proposed framework first employs the Depth-Anything encoder to extract the \textit{visual depth features} within traffic scenes, along with a “Faster R-CNN \cite{ren2016faster} + Spatiotemporal-GCN” and a pre-trained I3D \cite{carreira2017quo} encoder to capture \textit{visual interaction features} between agents and \textit{visual dynamic features} of the entire scenes. These three types of visual features (i.e., \textit{visual depth features}, \textit{visual interaction features}, and \textit{visual dynamic features}) jointly form a multi-dimensional visual feature for each video frame. After that, with each frame as a node and the multi-dimensional visual feature of each frame as the node feature, a frame graph is constructed. Finally, a graph convolution attention network \cite{brody2021attentive} is built that learns to focus on the early frames where the precursory visual features of accidents are prominent, so as to achieve accident anticipation as timely as possible.

It is worth noting that the primary innovation of this work lies in the integration of visual depth features in both \textbf{local} and \textbf{global} forms within our framework. Specifically, the \textit{visual depth features} are incorporated in \textbf{a local form} into the computation of \textit{visual interaction features}. This addresses the limitation of existing methods that only leverage appearance and 2D position to model interactions, which cannot accurately reflect the real spatial distance between agents. Additionally, the \textit{visual depth features} are also leveraged in \textbf{a global form} to construct the frame graph. This has the potential to enhance the model's holistic depth perception of the entire scene, rather than focusing on only the agents. We argue that this is important because some accidents may not occur between agents, but rather between an agent and the environment, such as a vehicle colliding with a curb.

To sum up, our main contributions are as follows:

\begin{itemize}
    \item We propose a novel and physical depth-aware learning framework for early accident anticipation. This framework fully utilizes the depth features extracted by the large model (i.e., Depth-Anything) to more accurately perceive the traffic agents and environments, thereby reducing the false alarms.
    \item We fully utilize three types of visual features from the dashcam that can reflect accident risks to achieve more holistic accident anticipation. These visual features are further integrated into a frame graph and fed into a graph attention network to adaptively capture the precursor features of accidents.
\end{itemize}

\section{RELATED WORKS}\label{sec2}

Early accident anticipation aims to estimate the probability of future accidents based on perceptual information from the current scene. Based on the differences in traffic scene modeling approaches, we categorize early accident prediction research into attention-based methods and graph convolution network (GCN) -based methods.
\subsection{Attention-based Methods}

Attention-based accident scene modeling methods \cite{karim2022dynamic,fatima2021global,chan2017anticipating,bao2020uncertainty,zeng2017agent,shah2018cadp,corcoran2019traffic,wang2020multitask,li2021abssnet,karim2022toward,dosovitskiy2020image,kang2022vision,karim2023attention} leverage the capability of attention mechanisms to dynamically focus on key features that may lead to traffic accidents, which have gained increasing attention in recent years. For example, Chan et al. \cite{chan2017anticipating} first introduced a Dynamic-Spatial-Attention (DSA) mechanism, which effectively addressed the limitation of earlier studies that overlooked dynamically changing features in video data and improved the accident prediction accuracy significantly. Zeng et al. \cite{zeng2017agent} proposed a novel soft attention mechanism using a Dynamic Parameter Prediction layer to model the relative spatial relationships and appearance couplings between agents and risk areas, which addressed the often-neglected impact of complex interactions among traffic participants on accident risk assessment. Shah et al. \cite{shah2018cadp} integrated contextual information into the Faster R-CNN model, focusing the model's attention on small-sized objects. Corcoran et al. \cite{corcoran2019traffic} emphasized the importance of temporal feature extraction for early traffic accident anticipation by proposing a dual-stream dynamic attention recurrent convolutional neural network framework. Wang et al. \cite{wang2020multitask} proposed a Multi-task Attention Network (MATNet) for lane detection, incorporating spatial and channel attention mechanisms to effectively enhance lane feature recognition. Li et al. \cite{li2021abssnet} focused on understanding traffic scenes, particularly on accurately segmenting different traffic participants in complex environments. They introduced an Attention-Based Spatial Segmentation Network (ABSSNet), which improved the accuracy and efficiency of segmenting important detection objects like vehicles and pedestrians in complex traffic scenes. Bao et al. \cite{bao2020uncertainty} combined attention mechanisms with deep reinforcement learning (DRL) and proposed two attention mechanisms (Bottom-Up \& Top-Down) that simulate the human visual system, enabling the model to dynamically adjust its focus on the environment and identify key factors that may lead to accidents. Karim \cite{karim2022toward} introduced a post-hoc attention mechanism Grad-CAM, to generate saliency maps, which visualize the regions of interest that the model focuses on when performing traffic accident prediction tasks. For early prediction of collisions between autonomous and non-autonomous vehicles, Fatima et al. \cite{fatima2021global} proposed a Feature Aggregation (FA) module that identifies abnormal events in video sequences by calculating the weighted sum of features of all objects in a given frame and refining each object's features. The Vision Transformer \cite{dosovitskiy2020image} (ViT) is a powerful image model introduced in recent years with self-attention mechanisms. Inspired by this, Kang et al. \cite{kang2022vision} proposed the Vision Transformer with Temporal Attention (ViT-TA). This transformer accurately classified critical situations around traffic accidents and highlighted key objects in accident conflict scenes based on attention maps, identifying potential causes of accidents. Besides integrating attention mechanisms into spatial scene modeling, Karim et al. \cite{karim2022dynamic} constructed a Dynamic Spatio-temporal Attention Network (DSTA), which incorporated a Dynamic-Temporal-Attention (DTA) module to track and analyze the development of key temporal features throughout the video sequence, thereby enabling early warnings of potential accidents. In their latest research, Karim et al. \cite{karim2023attention} developed an Attention-Guided Multi-stream Feature Fusion Network (AM-Net). In this network, the attention mechanism selectively weights features, determining which features were more critical for risk prediction and adjusting the feature fusion weights accordingly, which provided greater flexibility in handling complex and dynamic traffic environments.

\subsection{GCN-based Methods}

Spatial relationships between traffic agents are also important cues for early accident anticipation, as they often indicate potential interactions that may lead to collisions. As such, numerous studies \cite{malawade2022spatiotemporal,bao2020uncertainty,guo2019attention,diao2019dynamic,liu2020spatiotemporal,zhao2019t,wang2023gsc,sun2024maformer,thakur2024graph} have leveraged Graph Convolutional Networks (GCN) \cite{kipf2016semi} to intuitively model spatial relationships between traffic agents. For example, Guo et al. \cite{guo2019attention} proposed the Attention-based Spatio-Temporal Graph Convolutional Network (ASTGCN), which utilized GCN's ability to process data directly on the original graph-structured traffic network and captures the spatial features of different nodes within the traffic network effectively. Similarly, Diao et al. \cite{diao2019dynamic} introduced the Dynamic Spatio-Temporal Graph Convolutional Neural Network (DST-GCNN) to specifically capture the spatiotemporal dynamics of traffic data. Building on this, Liu et al. \cite{liu2020spatiotemporal} proposed a Spatio-Temporal Relationship Graph Convolutional Network (STR-GCN), which uses GCN to capture the spatial positional relationships of pedestrians within traffic scenes and their interactions with other traffic participants. Zhao et al. \cite{zhao2019t} combined the advantages of GCN's ability to capture spatial structural features in traffic networks with Recurrent Neural Networks (RNN) \cite{godard2017unsupervised} for analyzing temporal sequence data, creating a framework (T-GCN) capable of handling both spatial and temporal characteristics of traffic data. Considering the uncertainty in spatial relationships among traffic participants, Bao \cite{bao2020uncertainty} introduced uncertainty analysis into his research. This approach, which quantifies uncertainty within the GCN framework, adds a new dimension to accident prediction. Malawade et al. \cite{malawade2022spatiotemporal} enhanced the understanding of dynamic changes in traffic scenes by utilizing Graph Neural Networks and Long Short-Term Memory (LSTM) \cite{shi2015convolutional} to construct a spatiotemporal scene graph embedding model (SG2VEC). Wang et al. \cite{wang2023gsc} innovatively incorporated dynamic traffic elements (such as historical vehicle trajectories) into the construction of spatiotemporal graphs. By analyzing the complex interactions among these elements through GCN, they identified key features affecting accident collisions. Most recently, Thakur \cite{thakur2024graph} proposed a dual-layer nested graph architecture (GraphGraph), where the inner graph captures interactions among objects within video frames, and the outer graph analyzes spatiotemporal relationships between frames. This nested graph structure allows GraphGraph to effectively integrate global spatiotemporal information while capturing and understanding the complex dynamics among traffic elements over time.

\subsection{Limitations of existing methods}

Despite significant progress in accident anticipation, existing attention-based and graph convolution-based methods exhibit limitations in accurately capturing spatial relationships. They fail to incorporate depth features, hindering accurate 3D scene reconstruction from 2.5D video perspectives. This limitation leads to challenges in distinguishing collisions from mere overlap of traffic participants, resulting in false positive alerts. Furthermore, these methods overlook the impact of occluded key participants that are crucial for risk identification in complex scenarios, and tend to miss key features by focusing solely on either global information or local details.

\section{METHODOLOGY}\label{sec3}

This section discusses the details of our proposed model. The early accident prediction task has two main requirements: predicting as early as possible and accurately forecasting future accidents. For the first requirement, we divide a segment of dashcam video data into a sequence $X_{seq}\in(X_{1}X_{2},...,X_{N}).$ Assuming an accident occurs at frame \textit{a}, and a prediction is made at frame \textit{t}, we define the Time-to-Accident (TTA) as $\tau=a-t$, while $\mathit{t} < \mathit{a}$. Our goal is to maximize $\tau$ while maintaining prediction accuracy. For the second requirement, we demand that the prediction at frame \textit{t} yields the probability of an accident occurring, denoted as $p_{t}>$ \textit{threshold}, where \textit{threshold} is a predetermined threshold to ensure model accuracy. We aim to maintain high prediction accuracy while maximizing $p_{t}$, providing ample response time for drivers and advanced driver-assistance systems (ADAS).

\begin{figure*}[htb!]
  \centerline{\includegraphics[width=\textwidth]{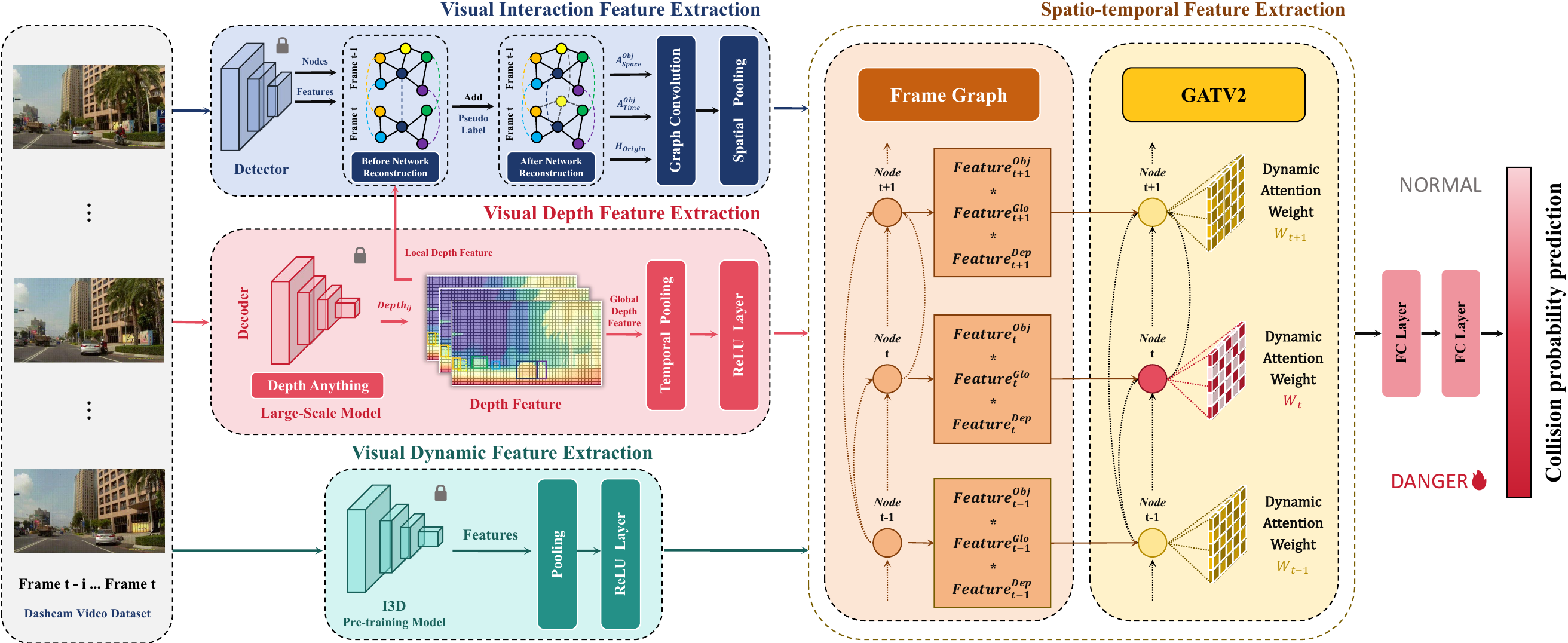}}
  \caption{ The overview of the proposed framework. Three types of visual features are obtained through pre-trained encoders. Visual depth feature represents the depth characteristics in each video frame, visual interaction feature captures the interaction relationships between traffic agents, and visual dynamic feature denotes the spatiotemporal dynamic changes in the video sequence. These visual features are then fused and passed to the graph attention layer and FC layers to generate the final prediction probability. The lock symbol indicates that the module is frozen during training.
  \label{fig2}}
  \end{figure*}

We propose a physical depth-aware early accident anticipation framework, as shown in Fig.\ref{fig2}. Our framework consists of four modules. 1) \textit{Visual Depth Feature Extraction Module}, which extracts depth features of traffic agents and images using the encoder of pre-trained Depth anything \cite{yang2024depth}. 2) \textit{Visual Interaction Feature Extraction Module}, which utilizes pre-trained object detectors to extract object features and labels, combining them with individual depth features to create a spatio-temporal object graph. 3) \textit{Visual Dynamic Feature Extraction Module}, which extracts spatiotemporal global features using a pre-trained I3D \cite{carreira2017quo} encoder and applies pooling to obtain the dynamic changes of the video sequence. 4) \textit{Spatio-Temporal Feature Extraction Module}, which fuses the three types of visual features and outputs the predicted probability of accident occurrence using graph attention and fully connected layers.

\subsection{Visual Depth Feature Extraction Module}

Local and global forms of visual depth features play a crucial role in understanding traffic accident scenes and early accident anticipation. As such, we utilize a foundational depth estimation model \textit{Depth Anything}, which is trained on a large-scale dataset comprising 1.5 million paired image-depth observations and 62 million unlabeled images, to extract both local and global depth features of the accident scenes. Specifically, the global depth features of the entire image are extracted using the Dense Prediction Transformer (DPT) encoder from \cite{yang2024depth} along with a pooling operation, and the local depth features are cropped from the global ones using a state-of-the-art object detector \cite{ren2016faster}.

\textbf{Algorithm 1} summarizes the process of depth features extraction. After extracting frames from the video sequence, the pre-trained detector \cite{ren2016faster} first produces the positions $X_{i,j}^L$ of agents in the current frame. The video sequence frame $X_{i}$ is then input into the pre-trained DPT encoder to obtain the global depth features $f_{Depth}\in\mathbb{R}^{B\times C\times H\times W}$, where the depth feature channel \textit{C} is 1. After that, the depth features list is filtered based on the positions of all bounding boxes in the current frame to extract the depth features $f_{i,j}^{depth}$ of each agent $A_{i,j}$ in the frame. To avoid issues of features dimension mismatch and high computational cost during multi-dimensional visual features fusion, the global depth features are adjusted to $f_{Depth}^{'}\in\mathbb{R}^{B\times(H\times W)}$ through a fully connected (FC) layer and a ReLU layer,
\begin{equation}
\begin{split}
f_{\text{Depth}} &= \mathrm{DPT}(X_{i}) \\
f_{\text{Depth}}^{'} &= \phi(f_{\text{Depth}})
\end{split}
\label{eq1}
\end{equation}
where $\phi$ represents the FC layer and ReLU layer utilized for dimensionality adjustment.

The \textit{Visual Depth Feature Extraction Module} effectively extracts both local and global depth features, which are then transmitted to the \textit{Visual Interaction Feature Extraction Module} and subsequently to the \textit{Spatio-Temporal Feature Extraction Module}.

\begin{algorithm}
\caption{The process of extracting visual depth features in both global and local forms}
\label{alg1}
\setlength{\algorithmicindent}{1.5em}
\renewcommand{\algorithmicrequire}{\textbf{Input:}}
\renewcommand{\algorithmicensure}{\textbf{Output:}}
\begin{algorithmic}[1]
\REQUIRE Sequence frame of $X$ dashcam video at each time step.
\ENSURE Local visual depth features $f_{i,j}^{depth}$ and global visual depth features $f_{Depth}^{'}$ at each frame.

\STATE Faster R-CNN detector $dec$.
\STATE Detected agents' location $X_{i,j}^L$.
\FOR{$X_{i} \in X$ \textbf{do}}
    \STATE$X_{i,j}^L = dec(X_i)$
    \STATE$f_{Depth}$ = DPT$(X_i)$
    \FOR{$j \in X_{i,j}^{L}$ \textbf{do}}
        \STATE $f_j^{depth} = f_{Depth}(X_{i,j}^L)$
        \IF{$f_{j}^{depth}$ \textbf{then}}
            \STATE Add $f_{j}^{depth}$ into $f_{i,j}^{depth}$ 
        \ENDIF
    \ENDFOR
    \STATE FC layer $f_{Depth}$ = $fc\_layer(f_{Depth})$
    \STATE ReLU layer $f_{Depth}^{'}$ = $ReLU(f_{Depth})$
\ENDFOR
\RETURN Local visual depth features $f_{i,j}^{depth}$ and global visual depth features $f_{Depth}^{'}$.
\end{algorithmic}
\end{algorithm}

\subsection{Visual Interaction Feature Extraction Module}

Road traffic accidents are often the result of a combination of factors, with the dynamic interactions between traffic agents playing a significant role. Our goal is to extract these interaction features, allowing the model to better focus on agent interactions and reduce background noise, thereby aiding in more accurate accident prediction. To achieve this, we propose to combine the local depth features extracted from the \textit{Visual Depth Feature Extraction Module} with the object features of the traffic agents. By doing so, we aim to analyze the interactions between traffic agents from a three-dimensional perspective, thereby identifying the critical factors that contribute to the occurrence of accidents. 

Modeling traffic scenes using graph-structured data can reorganize complex scenes and effectively capture the interaction features between objects \cite{bao2020uncertainty}. Specifically, the first step of the \textit{Visual Interaction Feature Extraction Module} is to process each sequence frame $X_{i}$ using an advanced object detector \cite{ren2016faster} to obtain the object features $f_{i,j}^{obj}\in\mathbb{R}^{m\times d_1}$ (where $d_1$ represents the dimension of object features) and labels $f_{i,j}^{label}\in\mathbb{R}^{m\times d_2}$ (where $d_2$ represents the dimension of labels) of the \textit{m} detected candidate objects in the current sequence frame. We use candidate objects as graph nodes and their positional relationships as edges to construct a complete traffic scene graph. Considering that depth features help better understand accident scenes, we also incorporate the local depth features $f_{i,j}^{depth}$ from the \textit{Visual Depth Feature Extraction Module} into the construction of graph nodes. However, high-dimensional node features can lead to significant computational costs in subsequent graph convolution calculations. Therefore, to maintain low-dimensional and representative features when constructing graph node features $N_{i}$, we introduce a fully connected (FC) layer to embed the candidate object's features, object labels, and depth features into the same low-dimensional space. Then, the embedded features are concatenated to enhance feature representation:
\begin{equation}
N_{i}=[\Phi(f_{i,j}^{obj}),\Phi(f_{i,j}^{label}),\Phi(f_{i,j}^{depth})]
\label{eq2}
\end{equation}
where $\Phi$ represents the fully-connected (FC) layer, and the operator [,] denotes feature dimension concatenation.

When constructing the scene graph edges, we consider that objects closer in distance are more likely to be involved in accidents in subsequent frames. Therefore, we define the spatial adjacency matrix $A_{mn}$ of the traffic scene graph as:
\begin{equation}
A_{mn}=\frac{\exp\{-d(c_m^i,c_n^i)\}}{\sum_{mn}\exp\{-d(c_m^i,c_n^i)\}}
\label{eq3}
\end{equation}
where $d(c_m^i,c_n^i)$ represents the Euclidean distance between the centers $c_m^i$ and $c_n^i$ of \textit{m}-th and \textit{n}-th object bounding boxes detected in the frame $X_i$, respectively. This means that as the distance between objects decreases, the graph convolution will apply greater weight in learning the interaction relationship between objects \textit{m} and \textit{n} Thus, the adjacency matrix defined by Eq. \eqref{eq3} has the advantage of suppressing the influence of irrelevant objects with particularly large pixel distances.

This spatial adjacency matrix $A_{mn}$ is constructed based on object detection in each frame. Existing methods typically use this representation, but they overlook the issue of perceiving actual physical distances inaccurately when using pixel space distances due to factors such as object occlusion. To address this, we propose to incorporate the features of occluded objects when constructing the traffic scene graph to make the spatial graph more complete. We introduce another reconstruction adjacency matrix $A_{rec}$, which incorporates the features of occluded objects predicted by the model into the traffic graph. When there is a mismatch in object labels between frames, we define the reconstruction adjacency matrix $A_{rec}$ as,
\begin{equation}
\begin{split}
&l_{rec}(i)=\theta(f_{i,j}^{obj},f_{i-t,j}^{obj}) \\
&A_{rec}=\begin{cases}l_{rec}(i),&\quad f_{i,j}^{label}\neq f_{i-1,j}^{label}\\0,&\quad\mathrm{otherwise}
\end{cases}
\end{split}
\label{eq4}
\end{equation}
where $f_{i,j}^{obj}$ and $f_{i-t,j}^{obj}$ represent the object features of frame $X_i$ and frame $X_{i-t}\left(t<i\right)$, respectively. $\theta$ represents the FC layer used for predicting the location information of occluded objects. During the reconstruction process, we use linear prediction to determine the current frame position of the occluded object based on the location of candidate objects from the present and previous frames, and fill in the occluded objects' attributes with the most recently observed features. The reconstructed spatial graph includes the feature information of occluded objects, preventing occlusion from affecting the spatio-temporal stability in subsequent training. Then, the constructed traffic scene graph is inputted into the Graph Convolutional Network (GCN) layer to extract visual interaction features,
\begin{equation}
\begin{split}
&f_{Int}=\mathrm{GCN}(N_{i},A_{mn}) \\
&f_{Int}^{rec}=\mathrm{GCN}(N_{i},A_{rec}) \\
&f_{Int}^{'}=[f_{Int},f_{Int}^{rec}]
\end{split}
\label{eq5}
\end{equation}

To adjust the feature dimensions and reduce computational load, we construct a Pooling layer and a ReLU layer after the GCN layer. Finally, the extracted visual interaction features are fed to the subsequent \textit{Spatio-temporal Feature Extraction Module} for feature fusion.

\subsection{Visual Dynamic Feature Extraction Module}

Signs of an impending accident are reflected not only in the interactions between traffic agents but also in the dynamic changes of the traffic environment. For example, a sudden narrowing of the lane can affect driver maneuvering, thereby increasing the risk of collisions. Unlike using pretrained models like Transformers \cite{sun2024maformer,ren2024unifying,jia2024finger} or ResNet \cite{qi2023resnet,shaheed2023efficientrmt,karthika2024improved} to extract global features, we propose to incorporate a pre-trained encoder of the I3D model \cite{carreira2017quo} to capture the global spatio-temporal dynamic features from accident video sequences. The I3D model, which is originally designed for action recognition in video data, can effectively process both spatial and temporal information, making it well-suited for identify critical dynamic features that are strongly correlated with accident occurrence.

In this module, the video frame sequence $X_{Seq}\in(X_{1},X_{2},...,X_{N})$ is first fed into the pre-trained encoder of an Inflated Inception-V1 \cite{carreira2017quo}. Subsequently, the features extracted by the Inception module are further aggregated in spatial dimension through pooling layer, and then an FC layer is used to reduce the feature dimensions. Finally, we obtain features $f_{Dyn}$ that contain the visual dynamic features of the entire video frames,
\begin{equation}
\begin{split}
&f_{I3D}=\mathrm{Inception}(X_{Seq}) \\
&f_{Dyn}=\phi(f_{I3D})
\end{split}
\label{eq6}
\end{equation}
where $\phi$ represents a pooling layer for feature aggregation followed by a FC layer for feature dimension reduction. The extracted visual dynamic features are then inputted to the subsequent \textit{Spatio-temporal Feature Extraction Module} for feature fusion.

\subsection{Spatio-Temporal Feature Extraction Module}

After extracting three types of visual features, they are fused to better identify the key cues that indicate early accidents. To this end, we construct a frame graph where each frame serves as a node, with its fused visual features representing the node attributes, and the temporal connections between frames acting as edges. For each frame $X_i$ in the video sequence, the visual depth features $f_{Depth}^{'}$, visual interaction features $f_{Int}^{'}$, and visual dynamic features $f_{Dyn}$ are concatenated to construct the node feature $f_{Node}$:
\begin{equation}
f_{Node}=[f_{Depth}^{'},f_{Int}^{'},f_{Dyn}]
\label{eq7}
\end{equation}

To prevent the leakage of feature information from future frames to past frames, which could compromise the integrity of accident prediction, we design the edges of the frame graph to be unidirectional, directed from past frames to future frames. Each ${Node}_i$ in the frame graph is connected to \textit{j} earlier nodes in the sequence via temporal edges (where $j<i$). We define the adjacency matrix of the frame graph $A_{Tem}$ as follows:
\begin{equation}
A_{Tem}(i,j)=\left\{\begin{matrix}1,&j<i\\0,&\text{otherwise}\end{matrix}\right.
\label{eq8}
\end{equation}

Then, the constructed frame graph is fed into the Graph Attention (GAT) \cite{brody2021attentive} layer, which learns to capture the importance of different time frames for predicting future accidents through the graph attention mechanism, i.e.,
\begin{equation}
f=\mathrm{GAT}(f_{Node},A_{Tem})
\label{eq9}
\end{equation}

Finally, we use the weighted matrix that reflects the importance of the spatio-temporal features captured by the graph attention mechanism and apply fully connected (FC) layers (denoted as $\delta$) to obtain the predicted probability $(P_1,P_2,\ldots,P_i)$ of an accident occurring at each time frame,
\begin{equation}
P_i=\delta(f_i)
\label{eq10}
\end{equation}

\subsection{Training objective}

In this framework, the Faster R-CNN detector, Depth Anything encoder, and I3D encoder are frozen to simplify computation and enhance prediction efficiency. During model training, we consider that the penalty for prediction failures should be more severe for time frames closer to the accident occurrence. To ensure fairness, we use exponential loss for training positive samples,
\begin{equation}
L_{Pos}(P_i)=\sum_{t=1}^T-e^{-\max(0,y-i)}\mathrm{log}(P_i^{pos})
\label{eq11}
\end{equation}
where \textit{y} denotes the accident occurrence time frame, \textit{i} refers to predicting at frame $X_i$, $P_i^{pos}$ represents the predicted probability of an accident occurring at frame $X_i$. For negative sample training, we use the standard cross-entropy loss, i.e.,
\begin{equation}
L_{Neg}(P_i)=\sum_{t=1}^T-\log(P_i^{neg})
\label{eq12}
\end{equation}
where $P_i^{neg}$ represents the predicted probability of non-accident occurring at frame $X_i$. The final training loss function for the model is the sum of the training losses for positive and negative samples, i.e.,
\begin{equation}
L_{pred}=\sum_{X\in Pos}L_{Pos}(P_{X_i})+\omega_1\sum_{X\in Neg}L_{Neg}(P_{X_i})
\label{eq13}
\end{equation}
where \textit{Pos} is the set of positive sample videos, \textit{Neg} is the set of negative sample videos, and $\omega_1$ is set to balance the impact of the imbalance between the number of positive and negative samples.

\section{EXPERIMENTS AND RESULTS}\label{sec4}

This section presents a comprehensive evaluation of the proposed physical depth-aware framework through extensive experiments conducted on multiple publicly available datasets. The experimental details, including dataset descriptions, hyperparameter settings, and evaluation metrics, as well as the evaluation results, are provided as follows.

\subsection{Datasets}
\begin{figure}[htbp]  
\centering  
\includegraphics[width=0.5\textwidth]{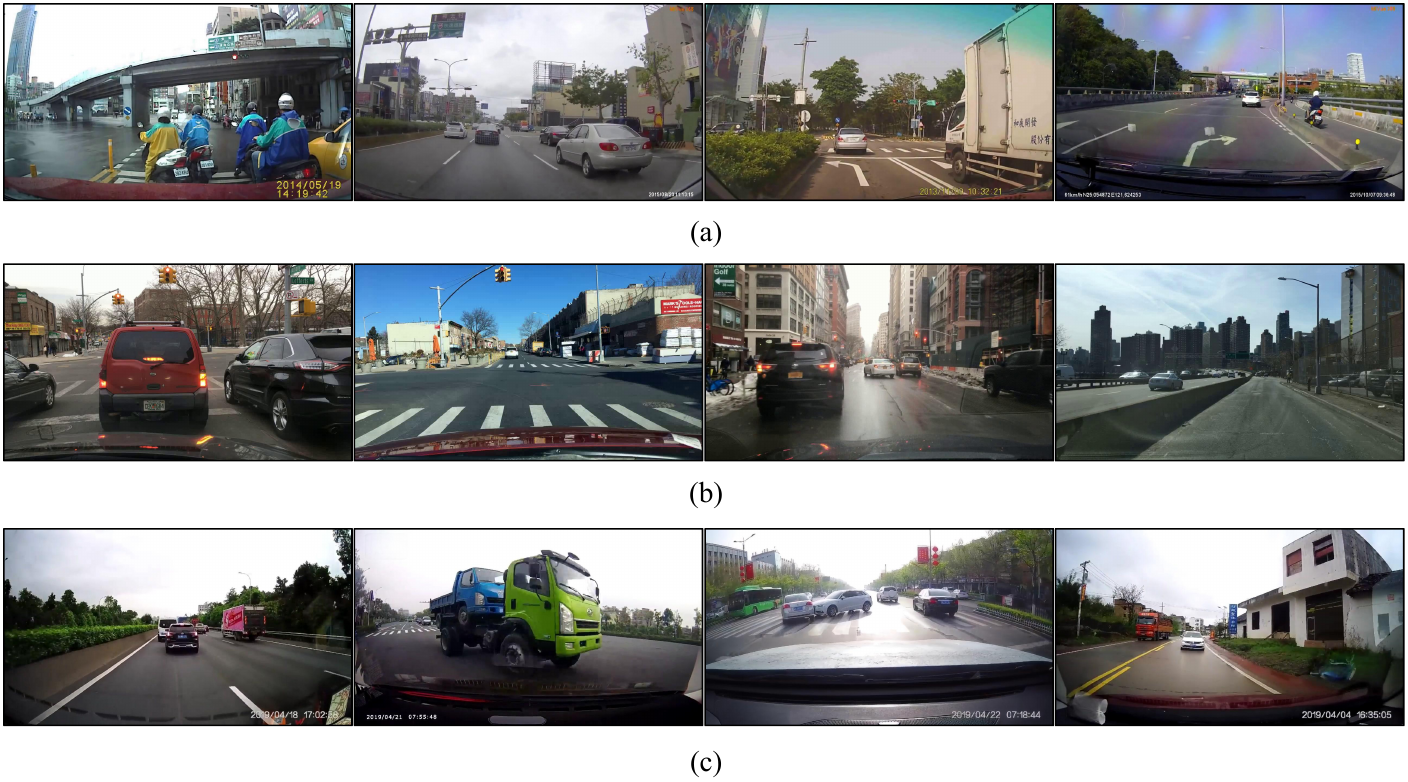}  
\caption{Public dashcam video datasets, (a) DAD, (b) CCD and (c) A3D.}  
\label{fig3}
\end{figure}

\subsubsection{DAD dataset}

The DAD \cite{chan2017anticipating} comprises 678 dashcam videos recorded in six major cities in Taiwan, with sample clips shown in Fig.\ref{fig3}(a). These videos have a frame rate of 20 frames per second (fps), with 58 videos used solely for training object detectors. The remaining 620 videos are sampled into 1,750 clips, each 100 frames (5 seconds) long. These clips include 620 positive videos containing accidents (starting at the 90th frame) and 1,130 negative videos without accidents, which are randomly allocated to the training and testing sets. The training set  consists of 455 positive videos and 829 negative videos, while the testing set contains 165 positive videos and 301 negative videos.

\subsubsection{CCD dataset}

The CCD \cite{bao2020uncertainty} comprises 4,500 dashcam videos annotated with various environmental attributes, with sample clips shown in Fig.\ref{fig3}(b). Each video has a frame rate of 10 frames per second (fps) and consists of 50 frames (5 seconds). These videos include 1,500 positive videos containing accidents (starting within the last 2 seconds) and 3,000 negative videos without accidents. Among these, 1,200 positive videos and 2,400 negative videos are used for the training set, while the remaining 300 positive videos and 600 negative videos are used for the testing set.

\subsubsection{A3D dataset}

The A3D \cite{yao2019unsupervised} comprises 1,500 dashcam videos captured in East Asia, with sample clips shown in Fig.\ref{fig3}(c). Similar to DAD, each video was sampled with 20 fps and consists of 100 frames, while the begining time of each accident was placed at the 80th frame. The dataset is randomly divided into 80\% for the training set and 20\% for the testing set. 

\subsection{The Detail of Implementation}

This study implemented the proposed method using Pytorch \cite{paszke2019pytorch}. Both training and testing were conducted on a workstation equipped with an NVIDIA A800 GPU (80G). To save computational cost, we used the pre-trained Depth anything \cite{yang2024depth} encoder to extract depth features, which had a dimension of 1,024 for both datasets. To ensure consistency in detected traffic agent feature attributes across different datasets, we employed the pre-trained Faster R-CNN \cite{ren2016faster} to obtain object feature information of candidate agents. Considering that over 90\% of frames in both datasets contained 19 or fewer traffic objects, we use the object detector to extract the top 19 objects of the highest detection score (\textit{m} = 19) to balance computational cost and prediction performance. The extracted object feature dimensions $d_1$ were 4,096, and the object label dimensions $d_2$ were 96, both adjusted to 512 via a fully connected embedding layer. Additionally, we used the pre-trained I3D \cite{carreira2017quo} encoder to extract global dynamic features, resulting in a feature dimension of 2,048. For hyperparameter tuning, we used ADAM \cite{kingma2014adam} as the optimizer and set the batch size to 10 for both DAD and CCD. The initial learning rate was set to 0.0001, and ReduceLROnPlateau \cite{sutskever2013importance} was utilized as the learning rate scheduler for stable model training.

\subsection{Evaluation Metrics}

The evaluation method for early accident prediction primarily relies on two metrics, i.e., average precision and time-to-accident. These metrics indicate the model's prediction accuracy and the lead time provided to drivers and ADAS.

\subsubsection{Average Precision}

This metric assesses the accuracy of the model's accident predictions. According to \cite{chan2017anticipating}, we assume that if the probability/confidence score of predicting a future accident at frame \textit{t} is higher than or equal to a threshold \textit{q}, an accident is expected to occur, otherwise, no accident is expected. Based on the classification of accident video samples (positive samples contain accidents), the model's predictions are categorized into True Positives (TP, positive sample correctly predicted as positive), False Positives (FP, negative sample incorrectly predicted as positive), False Negatives (FN, positive sample incorrectly predicted as negative), and True Negatives (TN, negative sample correctly predicted as negative). For each threshold \textit{q}, we calculate precision \textit{P} and recall \textit{R} using Eq. \eqref{eq14}. 
\begin{equation}
\begin{split}
P=\frac{{\mathrm{TP}}}{{\mathrm{TP}}+{\mathrm{FP}}} \\
R=\frac{{\mathrm{TP}}}{{\mathrm{TP}}+{\mathrm{FN}}}
\end{split}
\label{eq14}
\end{equation}

The precision value and recall value are changing when the classification threshold \textit{q} changes. We obtained multiple precision-recall pairs by setting different classification thresholds and constructed the corresponding precision-recall curve. The average precision (AP) is defined as the area under the precision-recall curve:
\begin{equation}
{\mathrm{AP}}=\int P_{R}dR
\label{eq15}
\end{equation}

When evaluating model performance, it is important to maintain both high precision and high recall. This prevents the model from overemphasizing precision while ignoring the potential severe consequences of false negatives in real-world traffic scenarios. Therefore, we followed recommendations from prior accident prediction studies \cite{chan2017anticipating,bao2020uncertainty,karim2022dynamic} and adopted the precision corresponding to 80\% recall, denoted by ${\mathrm{P}}_{80\%{\mathrm{R}}}$ as an additional metric to evaluate the predictive performance of the model.

\subsubsection{Time-to-Accident}

This metric measures the lead time of the model's accident predictions. When the model's prediction probability/confidence score for a positive sample video first exceeds the classification threshold, the time steps between the prediction frame $X_i$ and the actual accident occurrence frame $X_t$ are recorded as time-to-accident (TTA). We calculated the mean TTA (mTTA) for all positive sample videos and the TTA at 80\% recall (${\mathrm{TTA}}_{80\%{\mathrm{R}}}$) to evaluate the model's prediction lead time.

\begin{figure*}[htb!]
  \centerline{\includegraphics[width=\textwidth]{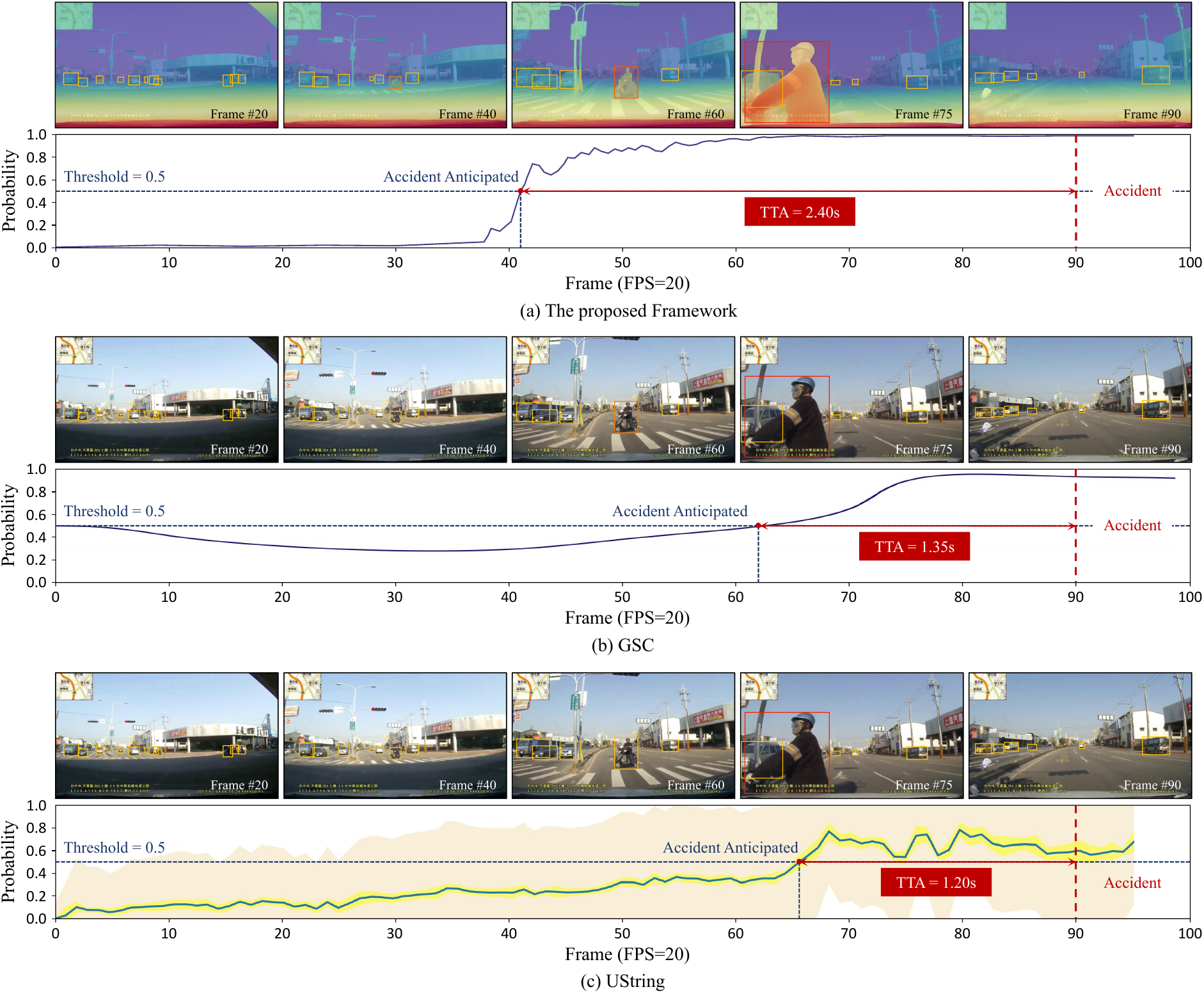}}
  \caption{ The visualization results of the proposed physical depth-aware framework, GSC, and UString on the same positive video sequence.
  \label{fig4}}
  \end{figure*}

\begin{table}[]
  \caption{Evaluation results on DAD, CCD and A3D datasets}
  \label{tab1}
  \centering
\renewcommand{\arraystretch}{1.3}
 \setlength{\tabcolsep}{0.8mm}{
\begin{tabular}{c|cc|cc|cc}
\toprule[0.8pt]
\multirow{2}{*}{Models}              & \multicolumn{2}{c|}{DAD\cite{chan2017anticipating}} & \multicolumn{2}{c|}{CCD\cite{bao2020uncertainty}} & \multicolumn{2}{c}{A3D\cite{yao2019unsupervised}} \\ \cmidrule{2-3} \cmidrule{4-5} \cmidrule{6-7}  
                                     & AP(\%)$\uparrow$         & mTTA(s)$\uparrow$        & AP(\%)$\uparrow$        & mTTA(s)$\uparrow$       & AP(\%)$\uparrow$        & mTTA(s)$\uparrow$        \\ \midrule \midrule
AdaLEA\cite{suzuki2018anticipating} & 52.3                     & 3.43                     & -                       & -                       & -                       & -                        \\
XAI\cite{karim2022toward}           & -                        & -                        & 94.0                    & 4.57                    & -                       & -                        \\
DSA\cite{chan2017anticipating}      & 48.1                     & 1.34                     & 99.6                    & 4.52                    & 92.3                    & 2.95                     \\
DSTA\cite{karim2022dynamic}         & 59.2                     & \textbf{3.66}                     & 99.6                    & \textbf{4.87}                    & 93.5                       & 2.87                        \\
UString\cite{bao2020uncertainty}    & 53.7                     & 3.53                     & 99.5                    & 4.74                    & 93.2                    & 3.24            \\
GSC\cite{wang2023gsc}               & 60.4                     & 2.55                     & 99.5                    & 3.58                    & 94.9                    & 2.62                     \\
Ours                                 & \textbf{64.6}            & 3.57            & \textbf{99.6}           & 4.82           & \textbf{95.7}           & \textbf{3.24}                     \\ \bottomrule[0.8pt]
\end{tabular}
}
\end{table}

\subsubsection{AUC}

The Area Under the Curve (AUC) represents the area under the Receiver Operating Characteristic (ROC) curve and serves as a metric to assess the model’s capability to distinguish between positive and negative samples. To construct the ROC curve, we first need to determine the False Positive Rate (FPR) and the True Positive Rate (TPR). The TPR is equivalent to recall, while the FPR represents the proportion of non-accident videos incorrectly predicted as positive samples to the total number of actual negative samples. The FPR can be calculated using the following formula:
\begin{equation}
\mathrm{FPR}=\frac{\mathrm{FP}}{\mathrm{FP}+\mathrm{TN}}
\label{eq16}
\end{equation}

After constructing the ROC curve, the AUC is calculated using the following mathematical formula:
\begin{equation}
\mathrm{AUC}=\int{R}(\mathrm{FPR}) d \mathrm{FPR}
\label{eq17}
\end{equation}

A larger AUC value indicates a stronger capability of the model to distinguish between positive and negative samples. An AUC value of 1 signifies that the model achieves perfect discrimination between positive and negative samples. An AUC value of 0 indicates that the model's predictions are entirely contrary to the actual labels. And an AUC value of 0.5 reflects that the model is effectively making random guesses regarding the sample classes.

\subsection{Evaluation Results}

In this section, we comprehensively evaluate the proposed physical depth-aware framework for early accident anticipation, including comparisons with state-of-the-art models, qualitative analysis of prediction results, and  further discuss the relationships between the evaluation metrics used to measure model performance.

\subsubsection{Comparison to the State-of-the-Art Models}

The proposed framework is compared with state-of-the-art models \cite{karim2022toward,karim2022dynamic,suzuki2018anticipating,wang2023gsc,chan2017anticipating,bao2020uncertainty} on the DAD, CCD and A3D datasets. Tab.\ref{tab1} presents the results of the comparative study. Notably, our framework achieved significant AP and mTTA on the challenging DAD dataset. We increased the AP from the highest 60.4\% achieved by GSC \cite{wang2023gsc} to 64.6\%, and extended the mTTA to 3.57 seconds. This indicates that the proposed framework significantly enhances early accident anticipation performance. Although the performance of the current state-of-the-art methods on the CCD and A3D datasets has nearly reached saturation, our framework still achieves 99.6\% AP with a 4.82-second mTTA on the CCD dataset, and 95.7\% AP with a 3.24-second mTTA on the A3D dataset, slightly surpassing the previously reported best-performing methods.

\begin{figure*}[htb!]
  \centerline{\includegraphics[width=\textwidth]{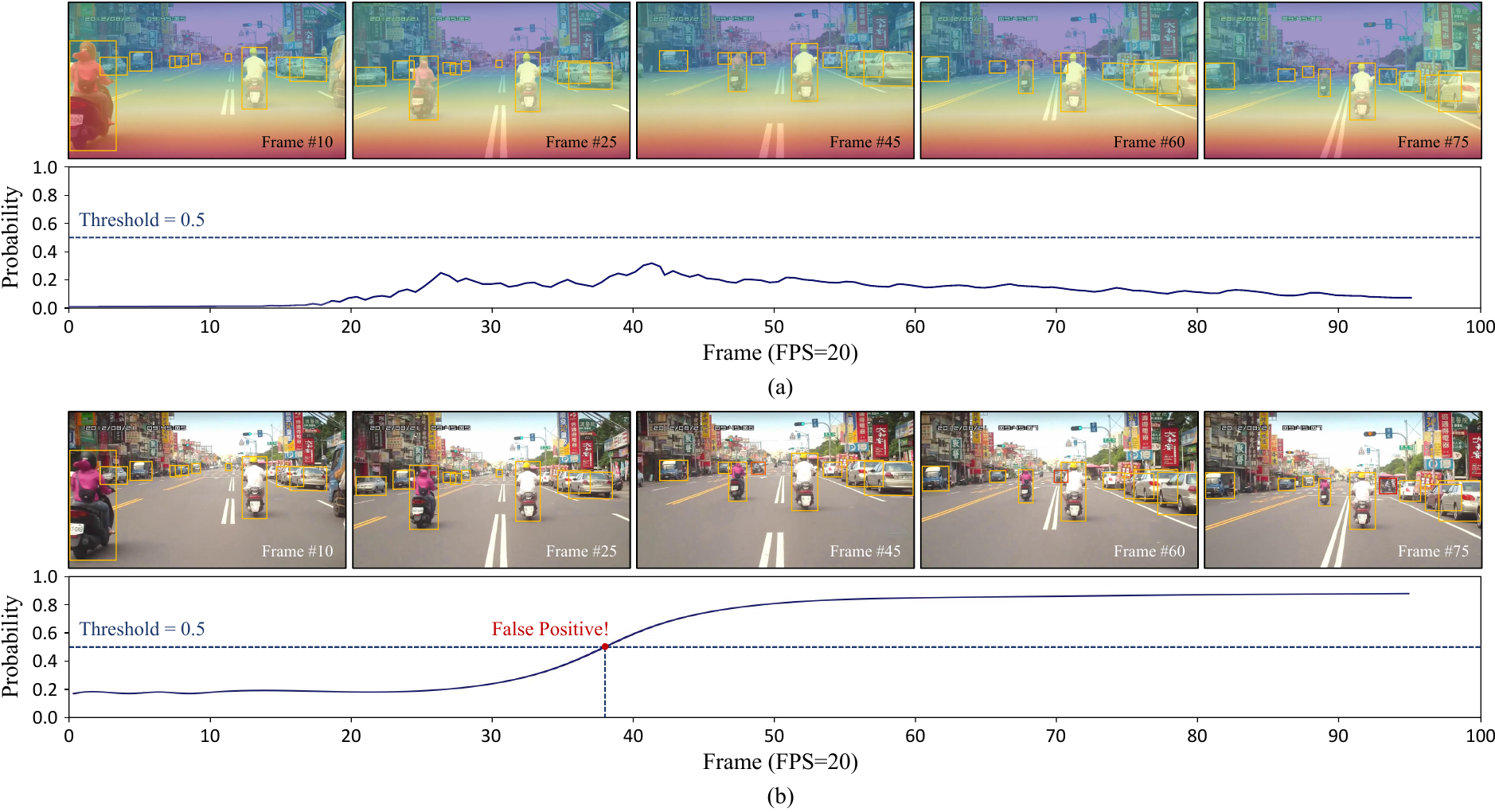}}
  \caption{ The visualization results of the proposed physical depth-aware framework and GSC on the same negative sample video sequence, where (a) shows the proposed framework identifying the video as True Negative. (b) shows GSC identifying the video as False Positive.
  \label{fig5}}
  \end{figure*}

\begin{table}[]
  \caption{Comparison of best AP of models on DAD }
  \label{tab2}
  \centering
\renewcommand{\arraystretch}{1.3}
 \setlength{\tabcolsep}{2mm}{
\begin{tabular}{cccccc}
\toprule[0.8pt]
Models  & AP(\%)$\uparrow$ & ${\mathrm{P}}_{80\%{\mathrm{R}}}$ & mTTA(s)       & ${\mathrm{TTA}}_{80\%{\mathrm{R}}}$ & AUC$\uparrow$ \\ \midrule[0.8pt]
UString\cite{bao2020uncertainty} & 72.2             & 49.3                              & 1.33          & 1.56                                & 0.69          \\
DSTA\cite{karim2022dynamic}    & 72.3             & 46.6                              & \textbf{1.50} & \textbf{1.81}                       & 0.70          \\
GSC\cite{wang2023gsc}     & 73.4             & 50.2                              & 1.25          & 1.47                                & 0.71          \\
Ours    & \textbf{75.8}    & \textbf{53.1}                     & 1.48          & 1.79                                & \textbf{0.73} \\ \bottomrule[0.8pt]
\end{tabular}
}
\end{table}

Tab.\ref{tab2} further compares the best AP achieved by the proposed framework and other well-performing models on the same dataset. Tab.\ref{tab1} has indicated that the models performed similarly on the CCD and A3D datasets. Therefore, we use the DAD dataset for comparison here. As shown in the table, our proposed framework achieved the best AP, improving it to 75.8\%. Meanwhile, the proposed model achieved the highest AUC, indicating that models with scene depth understanding can distinguish between accidents and non-accidents more accurately. Although the proposed model's mTTA and TTA were slightly surpassed by DSTA \cite{karim2022dynamic} by 0.02 seconds, false positives pose a greater threat in practical situations than a 0.02-second delay in alerts. Therefore, we believe that Tab.\ref{tab1} and Tab.\ref{tab2} demonstrate the competitiveness of our model.

To gain a deeper understanding of the proposed framework's performance, we conducted a case study on a representative positive sample video from the DAD dataset. We applied the proposed method, GSC\footnote{\url{https://github.com/ispc-lab/GSC}}\cite{wang2023gsc}, and UString\footnote{\url{https://github.com/Cogito2012/UString}}\cite{bao2020uncertainty} to a certain video and visualized their respective accident prediction performances, as depicted in Fig.\ref{fig4}. Specifically, Fig.\ref{fig4}(a) presents an exemplary instance of early accident anticipation by our proposed model, whereas (b) and (c) show the predictions by GSC and UString, respectively, for the same video sequence. In each example, the frame number is indicated at the bottom, with the prediction completion time marked by the intersection of the accident probability curve and the horizontal line indicates probability threshold 0.5. The ground truth (actual beginning time of accident) are labeled as the red dash line at frame \#90. In Fig.\ref{fig4}(a), the proposed framework reaches the threshold at frame \#41, yielding a 2.40-second TTA. Notably, the subsequent frames demonstrate the model's ability to accurately identify the collision risk caused by a car's sudden turn. For the same positive accident video, Figs.\ref{fig4}(b) and (c) illustrate that GSC and UString reach the threshold at frames \#62 and \#67, respectively, resulting in TTAs of 1.35 and 1.20 seconds. This confirms that the proposed method demonstrates superior early accident anticipation performance compared to other models in the same traffic scenarios, significantly aiding in alerting drivers/ADAS to prevent accidents. Furthermore, it highlights the significance of deep feature for the model's comprehensive understanding of traffic scenes and accurate accident prediction.

\subsubsection{Validation study}

To validate the significance of depth features in facilitating the model's in-depth understanding of traffic scenes, we revisited the False Positive example presented in \cite{wang2023gsc}. We conducted experiments on the proposed framework using the same negative sample video and visualized the accident prediction results, as shown in Fig.\ref{fig5}(a) and (b) illustrate the results of the proposed model and GSC on this negative sample video, respectively. The results in Fig.\ref{fig5}(a) indicate that our proposed model, by leveraging depth feature, gained a thorough comprehension of the traffic scene and correctly identified a true negative. In contrast, (b) shows that the GSC model, which solely relies on pixel distance between objects on a 2D plane, misinterpreted the true spatial position of the electric scooter, leading to a false positive prediction. The experimental results on the same negative sample video demonstrate the necessity of depth feature for constructing a complete spatial model of traffic scenes, effectively mitigating perspective distortion and similar visual phenomena that may lead to misleading accident predictions.

\begin{figure}[htbp]  
\centering  
\includegraphics[width=0.45\textwidth]{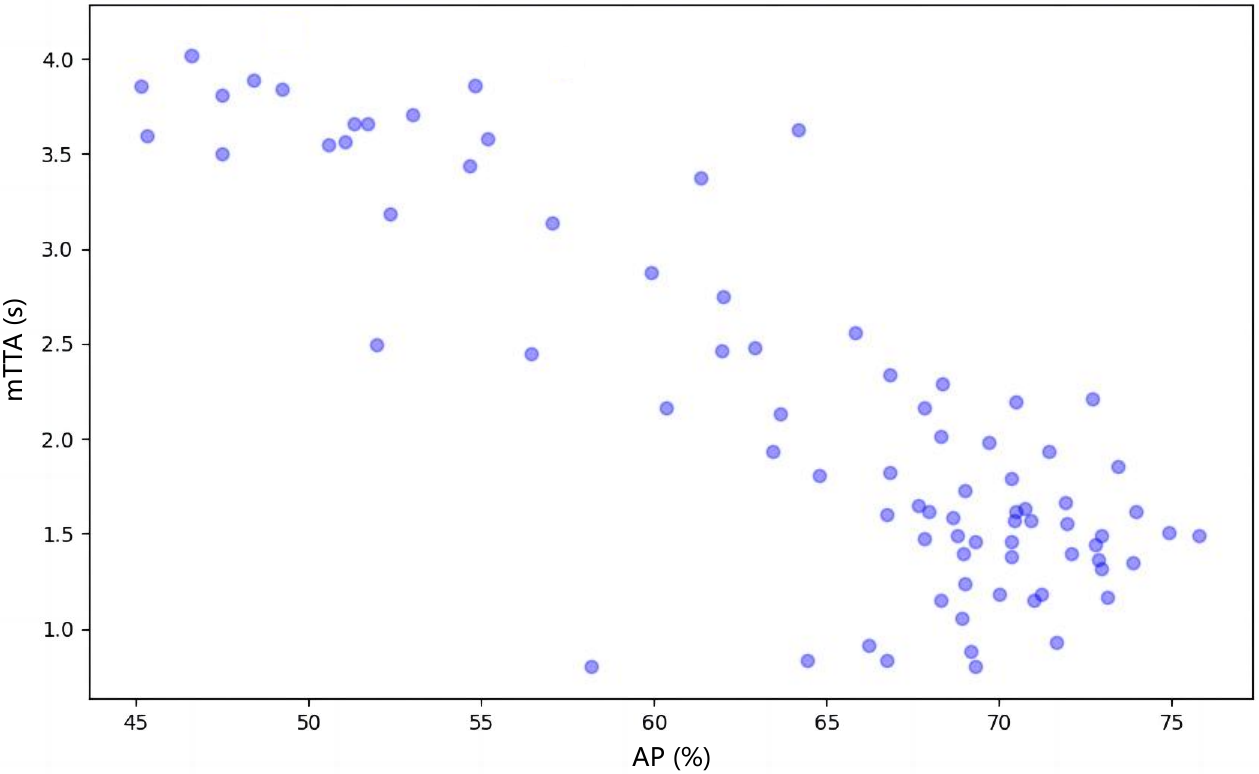}  
\caption{The performance of the proposed framework on DAD during training process.}  
\label{fig6}
\end{figure}

\subsubsection{Relationship between AP and mTTA metrics}

A learning behavior is observed from training the proposed framework, which seeks the highest AP and longest mTTA through multiple epochs while the model must balance prediction accuracy and anticipation time. According to Eq. \eqref{eq11}, the model is penalized more heavily for incorrect predictions on frames closer to the accident, encouraging earlier predictions. This results in higher average prediction accuracy but a shorter mTTA. Fig.\ref{fig6} illustrates the performance of the proposed model learning on the DAD. We only presented cases where the model achieves an AP greater than 45\% and an mTTA exceeding 0.8 seconds, as excessively low AP or mTTA values are not useful for practical early warnings. The figure shows a trade-off between the correctness and the earliness, where an increase in AP is accompanied by the shorten of mTTA. The proposed framework exhibited similar behavior when trained on the CCD and A3D datasets. To verify that the relationship between AP and mTTA is a general characteristic of early accident prediction tasks, we conducted additional experiments, recording AP and mTTA values for GSC \cite{wang2023gsc}, UString \cite{bao2020uncertainty} and DSTA \cite{karim2022dynamic} across different training epochs, as shown in Tab.\ref{tab3}. 

Tab.\ref{tab3} shows that as training epochs increase, the models' average prediction accuracy gradually increases while the mTTA value gradually shortens. This indicates that the model tends to sacrifice mTTA for higher prediction accuracy. Additionally, the experimental results demonstrate that under the same number of training epochs, the proposed model consistently outperforms the other models. It is important to note that mTTA should not be excessively long, as overly early predictions may generate false alarms, misleading normal driving. Therefore, striking a balance between AP and mTTA is crucial, based on the specific requirements of early accident anticipation in real traffic scenarios.

\begin{table}[]
  \caption{The relationship between AP and mTTA on DAD.}
  \label{tab3}
  \centering
\renewcommand{\arraystretch}{1.3}
 \setlength{\tabcolsep}{0.5mm}{
\begin{tabular}{c|cc|cc|cc|cc}
\toprule[0.8pt]
\multirow{2}{*}{\begin{tabular}[c]{@{}c@{}}Epoch\\ (x10)\end{tabular}} & \multicolumn{2}{c|}{Depth-aware} & \multicolumn{2}{c|}{GSC} & \multicolumn{2}{c|}{UString} & \multicolumn{2}{c}{DSTA} \\ \cmidrule{2-3} \cmidrule{4-5} \cmidrule{6-7} \cmidrule{8-9}  
                                                                       & AP(\%)         & mTTA(s)         & AP(\%)     & mTTA(s)     & AP(\%)       & mTTA(s)       & AP(\%)     & mTTA(s)     \\ \midrule \midrule
0                                                                      & 43.48          & 3.92            & 42.03      & 3.95        & 41.15        & 3.80          & 38.23      & 4.02        \\
5                                                                      & 45.21          & 3.85            & 43.52      & 3.83        & 44.09        & 3.69          & 40.81      & 3.94        \\
10                                                                     & 49.17          & 3.77            & 47.36      & 3.43        & 46.46        & 3.65          & 43.62      & 3.89        \\
15                                                                     & 55.67          & 3.61            & 52.10      & 3.27        & 50.79        & 3.48          & 47.55      & 3.72        \\
25                                                                     & 64.55          & 3.56            & 59.76      & 2.78        & 54.27        & 3.51          & 52.24      & 3.67        \\
50                                                                     & 68.38          & 2.37            & 63.46      & 1.38        & 57.91        & 2.26          & 56.88      & 2.42        \\ \bottomrule[0.8pt]
\end{tabular}
}
\end{table}

\subsubsection{Qualitative Results}

\begin{figure*}[htb!]
  \centerline{\includegraphics[width=\textwidth]{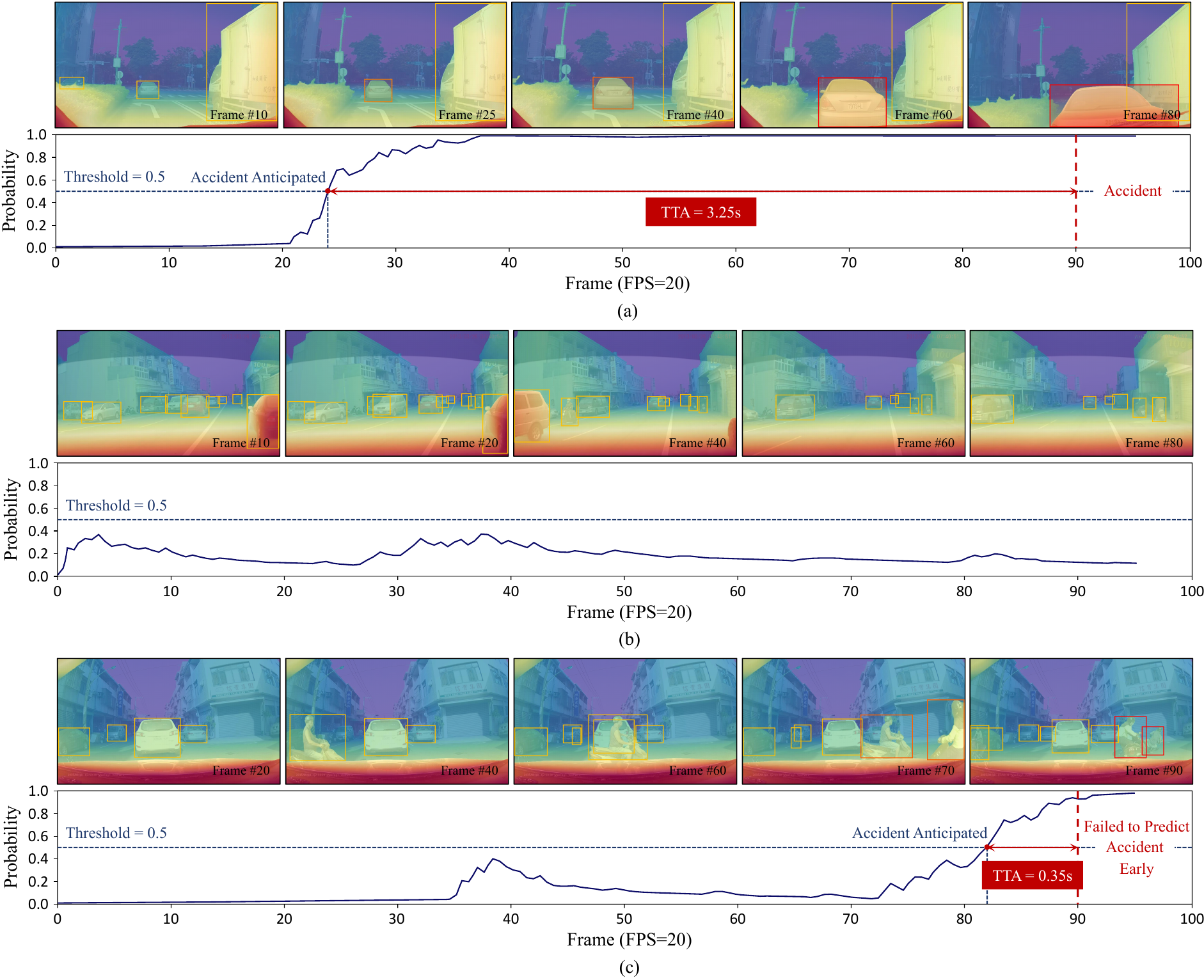}}
  \caption{ Examples of the proposed model’s early accident anticipation on DAD, where (a) shows a True Positive sample. (b) shows a True Negative sample. (c) illustrates a failure of the proposed model in early accident anticipation.
  \label{fig7}}
  \end{figure*}

In Fig.\ref{fig7}, we present the visualized prediction results of the proposed model on both positive and negative sample videos. Fig.\ref{fig7}(a) and (b) respectively show examples of successful predictions on positive and negative samples by the model. Fig.\ref{fig7}(c) depicts an example where the model successfully predicted an accident on a positive sample but with a short TTA. For illustration purposes, a threshold of 0.5 was used to trigger the completion of early accident prediction. In Fig.\ref{fig7}(a), the model predicted the probability of a future accident reaching the threshold at frame \#24, successfully completing the early accident prediction and generating a TTA of 3.25 seconds. Subsequent frames show that the model correctly identified the risk of an accident caused by an emergency lane change. In Fig.\ref{fig7}(b), despite vehicle occlusion, the model used detailed depth information of agents to construct a complete traffic scene, keeping the prediction probability below 50\% and avoiding misguidance. In Fig.\ref{fig7}(c), an electric scooter rapidly entered the traffic scene ahead of the ego vehicle at frame \#70. The abrupt change caused the model to reach the threshold at frame \#82, resulting in a TTA of only 0.35 seconds, which was insufficient for driver/ADAS to react. This failure in early accident prediction demonstrates that when the accident subject appears in the ego vehicle's field of view for a short duration, it can be difficult for the network to capture the key visual features leading to the accident.

\begin{table}[]
  \caption{Ablation studies results on DAD dataset}
  \label{tab4}
  \centering
\renewcommand{\arraystretch}{1.3}
 \setlength{\tabcolsep}{1.5mm}{
\begin{tabular}{c|c|c|cc|c|cc}
\toprule[0.8pt]
\rowcolor[HTML]{FFFFFF} 
\cellcolor[HTML]{FFFFFF}                           & \cellcolor[HTML]{FFFFFF}                       & \cellcolor[HTML]{FFFFFF}                      & \multicolumn{2}{c|}{\cellcolor[HTML]{FFFFFF}IEM}                         & \cellcolor[HTML]{FFFFFF}                      & \multicolumn{2}{c}{\cellcolor[HTML]{FFFFFF}Metrics}           \\ \cmidrule{4-5} \cmidrule{7-8} 
\rowcolor[HTML]{FFFFFF} 
\multirow{-2}{*}{\cellcolor[HTML]{FFFFFF}Variants} & \multirow{-2}{*}{\cellcolor[HTML]{FFFFFF}STFM} & \multirow{-2}{*}{\cellcolor[HTML]{FFFFFF}DEM} & \multicolumn{1}{c|}{\cellcolor[HTML]{FFFFFF}$A_{mn}$}     & $A_{rec}$    & \multirow{-2}{*}{\cellcolor[HTML]{FFFFFF}DYM} & \multicolumn{1}{c|}{\cellcolor[HTML]{FFFFFF}AP(\%)} & mTTA(s) \\ \midrule \midrule
\rowcolor[HTML]{EFEFEF} 
(1)                                                & $\checkmark$                                   & $\checkmark$                                  & \multicolumn{1}{c|}{\cellcolor[HTML]{EFEFEF}$\checkmark$} & $\checkmark$ & $\checkmark$                                  & \multicolumn{1}{c|}{\cellcolor[HTML]{EFEFEF}64.6}   & 3.78    \\
\rowcolor[HTML]{FFFFFF} 
(2)                                                & $\checkmark$                                   & $\checkmark$                                  & \multicolumn{1}{c|}{\cellcolor[HTML]{FFFFFF}$\checkmark$} &              & $\checkmark$                                  & \multicolumn{1}{c|}{\cellcolor[HTML]{FFFFFF}60.8}   & 3.66    \\
\rowcolor[HTML]{FFFFFF} 
(3)                                                & $\checkmark$                                   &                                               & \multicolumn{1}{c|}{\cellcolor[HTML]{FFFFFF}$\checkmark$} & $\checkmark$ & $\checkmark$                                  & \multicolumn{1}{c|}{\cellcolor[HTML]{FFFFFF}53.2}   & 3.74    \\
\rowcolor[HTML]{FFFFFF} 
(4)                                                & $\checkmark$                                   & $\checkmark$                                  & \multicolumn{1}{c|}{\cellcolor[HTML]{FFFFFF}}             &              & $\checkmark$                                  & \multicolumn{1}{c|}{\cellcolor[HTML]{FFFFFF}56.8}   & 3.34    \\
\rowcolor[HTML]{FFFFFF} 
(5)                                                & $\checkmark$                                   & $\checkmark$                                  & \multicolumn{1}{c|}{\cellcolor[HTML]{FFFFFF}$\checkmark$} & $\checkmark$ &                                               & \multicolumn{1}{c|}{\cellcolor[HTML]{FFFFFF}62.0}   & 2.65    \\
\rowcolor[HTML]{FFFFFF} 
(6)                                                & $\checkmark$                                   &                                               & \multicolumn{1}{c|}{\cellcolor[HTML]{FFFFFF}}             &              & $\checkmark$                                  & \multicolumn{1}{c|}{\cellcolor[HTML]{FFFFFF}42.7}   & 2.97    \\
\rowcolor[HTML]{FFFFFF} 
(7)                                                & $\checkmark$                                   &                                               & \multicolumn{1}{c|}{\cellcolor[HTML]{FFFFFF}$\checkmark$} & $\checkmark$ &                                               & \multicolumn{1}{c|}{\cellcolor[HTML]{FFFFFF}52.4}   & 2.65    \\ \midrule
\end{tabular}
}
\end{table}

\subsection{Ablation Study}

In this section, to verify the usefulness of different components, we designed six model variants for ablation experiments. By removing or replacing key components and recording the decline in model performance, we measured each component's contribution. Tab.\ref{tab4} summarizes the AP and mTTA of different model variants on the DAD dataset. Note that STFM, DEM, IEM, and DYM represent the \textit{Spatio-Temporal Feature Extraction Module}, \textit{Visual Depth Feature Extraction Module},\textit{ Visual Interaction Feature Extraction Module}, and \textit{Visual Dynamic Feature Extraction Module}, respectively. Additionally, $A_{mn}$ and $A_{rec}$ denote the spatial adjacency matrix and reconstruction adjacency matrix of IEM. Variant 1 is the network with all components of the proposed framework in place, achieving the highest AP and longest mTTA in the experiment. By comparing Variant 1 and Variant 3, we can see that the DEM component contributes the most to prediction accuracy, with an approximate 20\% gain, verifying the necessity of depth feature information for early accident anticipation. Similar observations can be seen from the comparison of Variant 7 and Variant 5 results and the comparison of Variant 6 and Variant 4 results. Furthermore, Variant 2 is a variant where the DEM component does not reconstruct occluded agents during spatial modeling, reducing prediction accuracy from 64.6\% to 60.8\%, verifying the robustness of the reconstructed adjacency matrix in improving model performance. The results of Variant 4 and Variant 5 demonstrate that the IEM and DYM components also effectively improve model performance, with the DYM component increasing the model's mTTA from 2.65 seconds to 3.78 seconds, proving the importance of incorporating global spatiotemporal dynamic features. The result of Variant 6 shows that the variant composed only of STFM and DYM achieved a maximum mTTA of 2.97 seconds, but the AP at this time was only 42.7\%, the worst prediction accuracy among all variants, indirectly confirming the contribution of DEM and IEM components to accident prediction accuracy. Variant 7, based on Variant 3, removes the DYM component, reducing the model's mTTA from 3.74 seconds to 2.65 seconds, further illustrating that the DYM component provides more response time for drivers/ADAS, effectively enhancing driving safety.

\section{CONCLUSION}\label{sec5}

This paper tackles the challenge of early accident anticipation in traffic scenarios being misled by perspective distortion, aiming to enable the model to understand complex traffic scenes from a three-dimensional perspective and avoid being misled by two-dimensional pixel distances. To this end, we propose a physical depth-aware early accident anticipation framework to handle this task. Specifically, the proposed framework leverages depth feature information extracted by a depth estimation large model encoder and integrates it with other crucial multi-dimensional visual features from dashcam videos, inputting these into a graph attention network to effectively capture key precursor features leading to accidents. We conducted qualitative and quantitative experiments on the DAD, CCD and A3D datasets to evaluate the robust performance of the proposed method in accident prediction tasks. Ablation experiments on different component variants further demonstrate the effectiveness of the proposed components.We hope that the success of using depth features in this study can provide more insights for future research on early accident anticipation.


%








\bibliographystyle{IEEEtran}
\bibliography{Bibliography}

\begin{IEEEbiography}[{\includegraphics[width=1in,height=1.25in,clip,keepaspectratio]{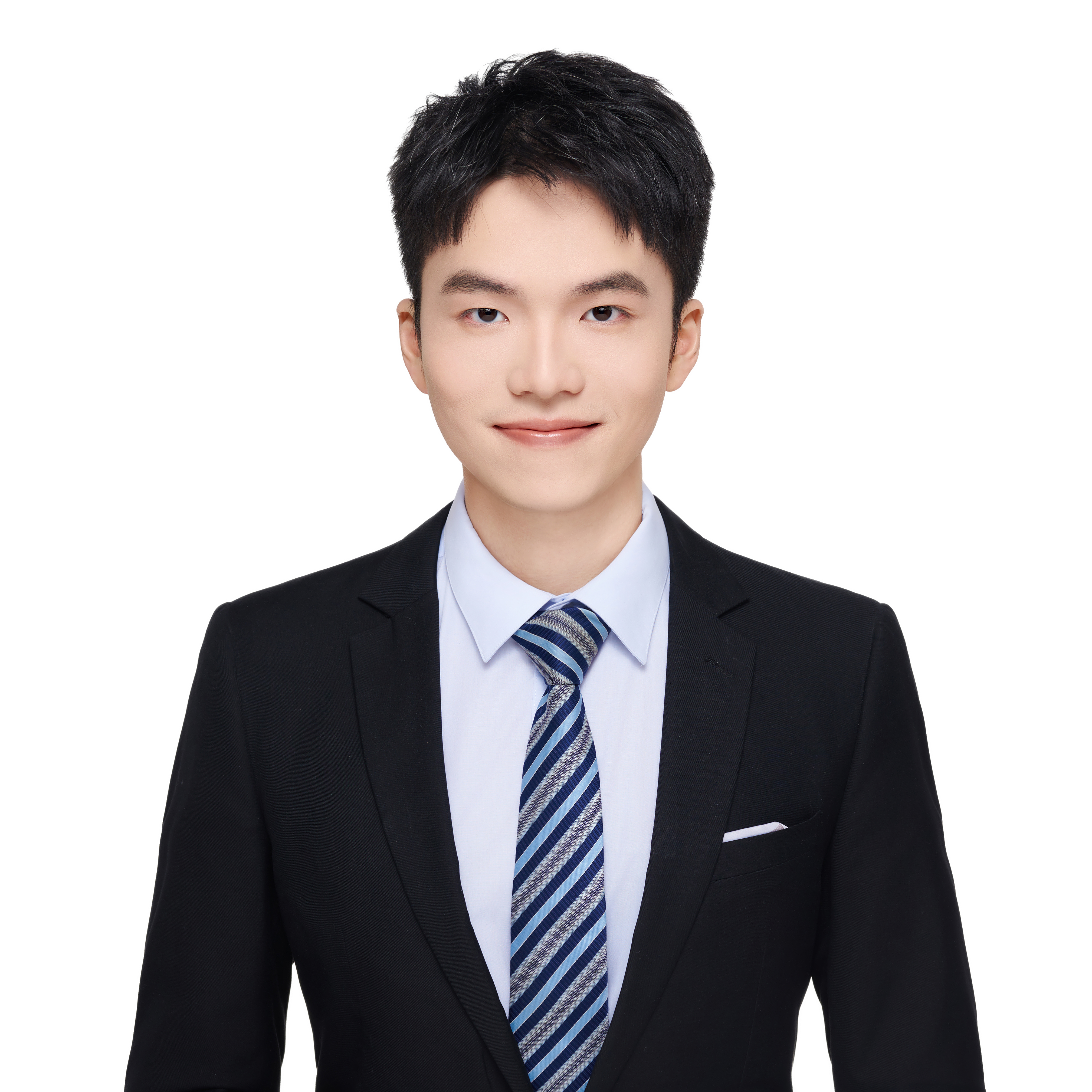}}]{Hongpu Huang} received the B.S. degree in the School of Transportation and Logistics, Southwest  Jiaotong University. He is currently pursuing the master degree with the School of Transportation, Southeast University. His researches focus on the application of deep learning and traffic prediction, especially on group activity recognition and accident anticipation. 
\end{IEEEbiography}

\vspace{-0.2in}

\begin{IEEEbiography}[{\includegraphics[width=1in,height=1.25in,clip,keepaspectratio]{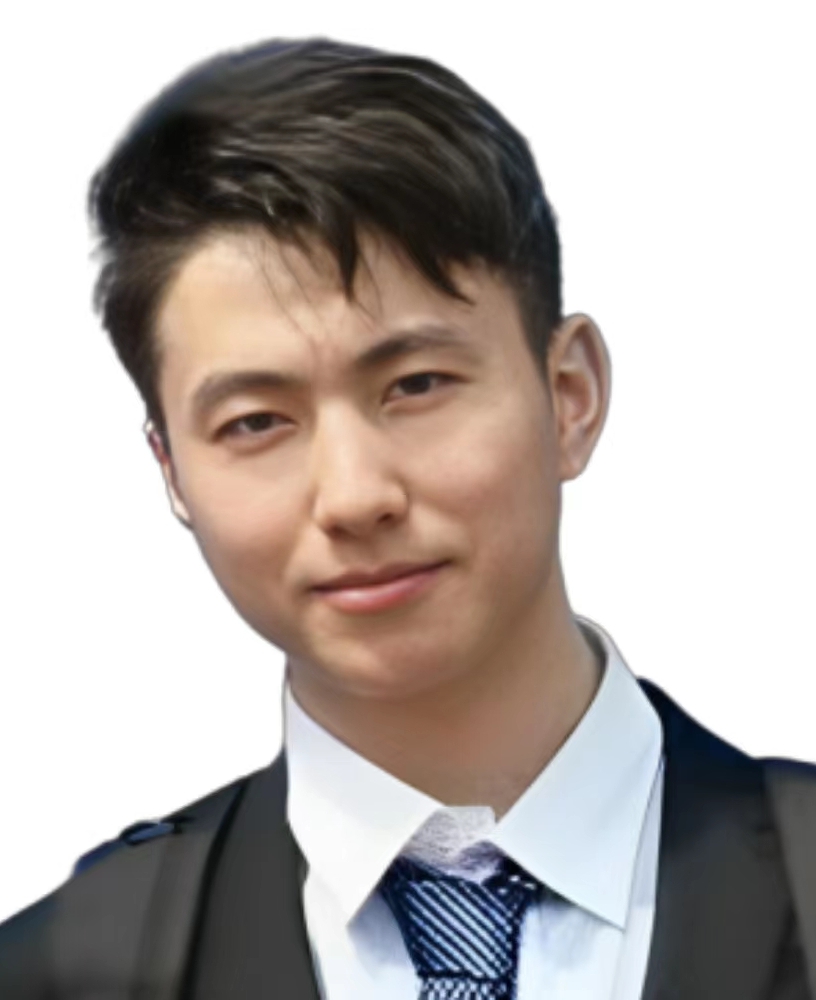}}]{Wei Zhou} received the B.S. degree from the School of Automation, Nanjing University of Science and Technology, and the Ph.D. degree from the School of Transportation, Southeast University. He is currently a Postdoctoral Research Fellow with the Department of Industrial and Systems Engineering, The Hong Kong Polytechnic University. His research focuses on vehicle detection, abnormal event detection, and large models. He has authored 12 articles (10 as first author, 2 as corresponding author) in JCR Q1 journals, including six in \textit{IEEE Transactions on Intelligent Transportation Systems}. He serves as an editorial board member for \textit{Digital Transportation and Safety} and \textit{Journal of Transportation Engineering and Information}, and reviews for leading journals including \textit{IEEE TITS/TIV/TII} and \textit{AAP}.
\end{IEEEbiography}

\vspace{-0.2in}

\begin{IEEEbiography}[{\includegraphics[width=1in,height=1.25in,clip,keepaspectratio]{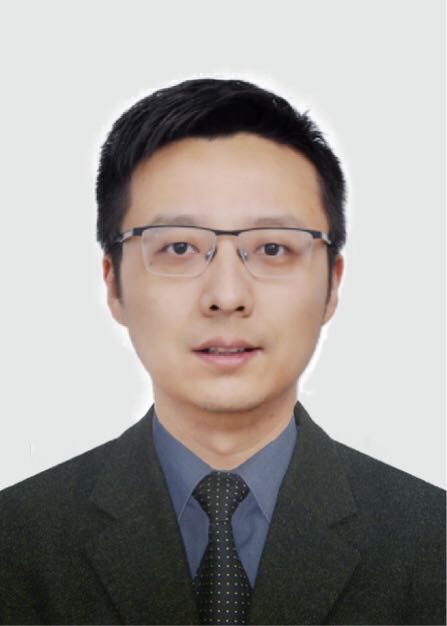}}]{Chen Wang}
  received the Ph.D. degree in civil and environmental engineering from the University of Kentucky, Lexington, KY, USA, in 2012. He is currently a Professor with the Intelligent Transportation Research Center, Southeast University, China. His main research interests include traffic big data analysis, connected and automated vehicles robust control, traffic safety, and microscopic traffic simulation. He serves as an Associate Editor for IEEE T-ITS and IEEE OJ-ITS.
\end{IEEEbiography}









\end{document}